\pdfoutput=1

\documentclass[11pt]{article}

\usepackage{authblk}

\usepackage{acl}

\usepackage{times}
\usepackage{latexsym}
\usepackage{graphicx}
\usepackage{tabularx}
\usepackage{booktabs}
\usepackage{amssymb}
\usepackage{multirow}

\usepackage[T1]{fontenc}

\usepackage[utf8]{inputenc}

\usepackage{microtype}

\usepackage{lipsum}

\newcommand\blfootnote[1]{%
  \begingroup
  \renewcommand\thefootnote{}\footnote{#1}%
  \addtocounter{footnote}{-1}%
  \endgroup
}

\title{Ensembling and Knowledge Distilling of Large Sequence Taggers for Grammatical Error Correction}

\author{Maksym Tarnavskyi$\dagger$\thanks{\ This research was performed during Maksym Tarnavskyi's work on Ms.Sc. thesis at Ukrainian Catholic University \cite{tar2021}.}}
\author[$\ddagger$]{Artem Chernodub}
\author[$\ddagger$]{Kostiantyn Omelianchuk}
\affil[$\dagger$]{Ukrainian Catholic University, Faculty of Applied Sciences, \texttt{tarnavskyi@ucu.edu.ua}}
\affil[$\ddagger$]{Grammarly, \texttt{firstname.lastname@grammarly.com}}

\begin{document}

\maketitle
\begin{abstract}
In this paper, we investigate improvements to the GEC sequence tagging architecture with a focus on ensembling of recent cutting-edge Transformer-based encoders in Large configurations. We encourage ensembling models by majority votes on span-level edits because this approach is tolerant to the model architecture and vocabulary size. Our best ensemble achieves a new SOTA result with an $F_{0.5}$ score of 76.05 on BEA-2019 (test), even without pre-training on synthetic datasets. In addition, we perform knowledge distillation with a trained ensemble to generate new synthetic training datasets, "Troy-Blogs" and "Troy-1BW". Our best single sequence tagging model that is pretrained on the generated Troy- datasets in combination with the publicly available synthetic PIE dataset achieves a near-SOTA\footnote{To the best of our knowledge, our best single model gives way only to much heavier T5 model \cite{rothe2021a}.} result with an $F_{0.5}$ score of 73.21 on BEA-2019 (test). The code, datasets, and trained models are publicly available.\footnote{\url{https://github.com/MaksTarnavskyi/gector-large}}
\end{abstract}

\section{Introduction}

The purpose of the Grammatical Error Correction (GEC) task is to correct grammatical errors in natural texts. This includes correcting errors in spelling, punctuation, grammar, morphology, word choice, and others. An intelligent GEC system receives text containing mistakes and produces its corrected version. The GEC task is complicated and challenging: the accuracy of edits, inference speed, and memory limitations are topics of intensive research.

Currently, Machine Translation (MT) is the mainstream approach for GEC. In this setting, errorful sentences correspond to the source language, and error-free sentences correspond to the target language. Early GEC-MT methods leveraged  phrase-based statistical machine translation (PBSMT) \cite{yuan-felice-2013-constrained}. Then this approach rapidly evolved to seq2seq Neural Machine Translation (NMT) based on gated recurrent neural networks \cite{yuan2016grammatical} and recent powerful Transformer-based seq2seq models. Transformer-based models autoregressively capture the full dependency among output tokens; however, inference can be slow due to sequential decoding. \citet{grundkiewicz-etal-2019-neural} leveraged a Transformer model  \cite{vaswani2017attention} that was pre-trained on synthetic GEC data and right-to-left re-ranking for ensemble. \citet{kaneko2020encoder} adopted several strategies of BERT \cite{devlin2018bert} usage for GEC. Recently, \citet{rothe2021a} built their system on top of T5 \cite{xue-etal-2021-mt5}, a xxl version of the T5 Transformer encoder-decoder model and reached new state-of-the-art results (11B parameters). 

While still not as widespread as MT, the sequence tagging approach for GEC, which generates a sequence of text edit operations encoded by tags for errorful input text is becoming more common. LaserTagger \cite{malmi-etal-2019-encode} is a sequence tagging model that casts text generation as a text editing task.  Corrected texts are reconstructed from the inputs using three main edit operations: keeping a token, deleting it, and adding a phrase before the token. LaserTagger combines a BERT encoder with an autoregressive Transformer decoder, which predicts edit operations. The Parallel Iterative Edit (PIE) model \cite{awasthi2019parallel} does parallel decoding, achieving quality that is competitive with the seq2seq models.\footnote{\url{http://nlpprogress.com/english/grammatical_error_correction}} It predicts edits instead of tokens and iteratively refines predictions to capture dependencies.\blfootnote{The paper has been accepted for publication at 60th Annual Meeting of the Association for Computational Linguistics (ACL 2022).} A similar approach is presented in \cite{omelianchuk2020gector}. The GECToR system achieves competitive results using various Transformers as an encoder; and linear layers with softmax for tag prediction and error detection. By replacing an autoregressive decoder with linear output layers, it’s also potentially several times faster than seq2seq systems.

Today, the generation of synthetic data is becoming significant for most GEC models. Natural languages are rich, and their grammars contain many rules and exceptions; therefore, professional linguists are often utilized to annotate high-quality corpora for further training ML-based systems mostly in a supervised manner \cite{dahlmeier2013building}, \cite{bryant2019the}. However, human annotation is expensive, so researchers are working on methods for augmentation of training data, synthetic data generation, and strategies for its efficient usage \cite{lichtarge2019corpora}, \cite{kiyono2019an}, \cite{stahlberg2021synthetic}. The majority of GEC systems today use synthetic data to pre-train Transformer-based components of their models.

In this work, we are focusing on exploring sequence tagging models and their ensembles. Although most of our developments may eventually be applied to other languages, we work with English only in this study. Being a resource-rich language, English is a highly competitive area for the GEC task$^3$.


\section{Base System Overview}
\subsection{GECToR architecture} 
Our tagging models are inherited from GECToR \cite{omelianchuk2020gector}. To date, GECToR shows near-SOTA results on CoNLL-2014 and BEA-2019 benchmarks.$^3$ It is based on AllenNLP \cite{Gardner2017AllenNLP} and  HuggingFace Transformers \cite{wolf2019huggingface}, and its source code is freely available.\footnote{\url{https://github.com/grammarly/gector}}

GECToR is a sequence tagging model that contains a Transformer-based encoder stacked with two output linear layers that are responsible for error detection and error correction. The model is trained with a cross-entropy loss function to produce tags that encode token-level edits. Then iterative postprocessing is performed. GECToR predicts the tag-encoded transformations for each token in the input sequence; it can then apply these transformations to get the modified output sequence. 

\begin{figure}[!h]
\centering
\includegraphics[width=1.0\columnwidth]{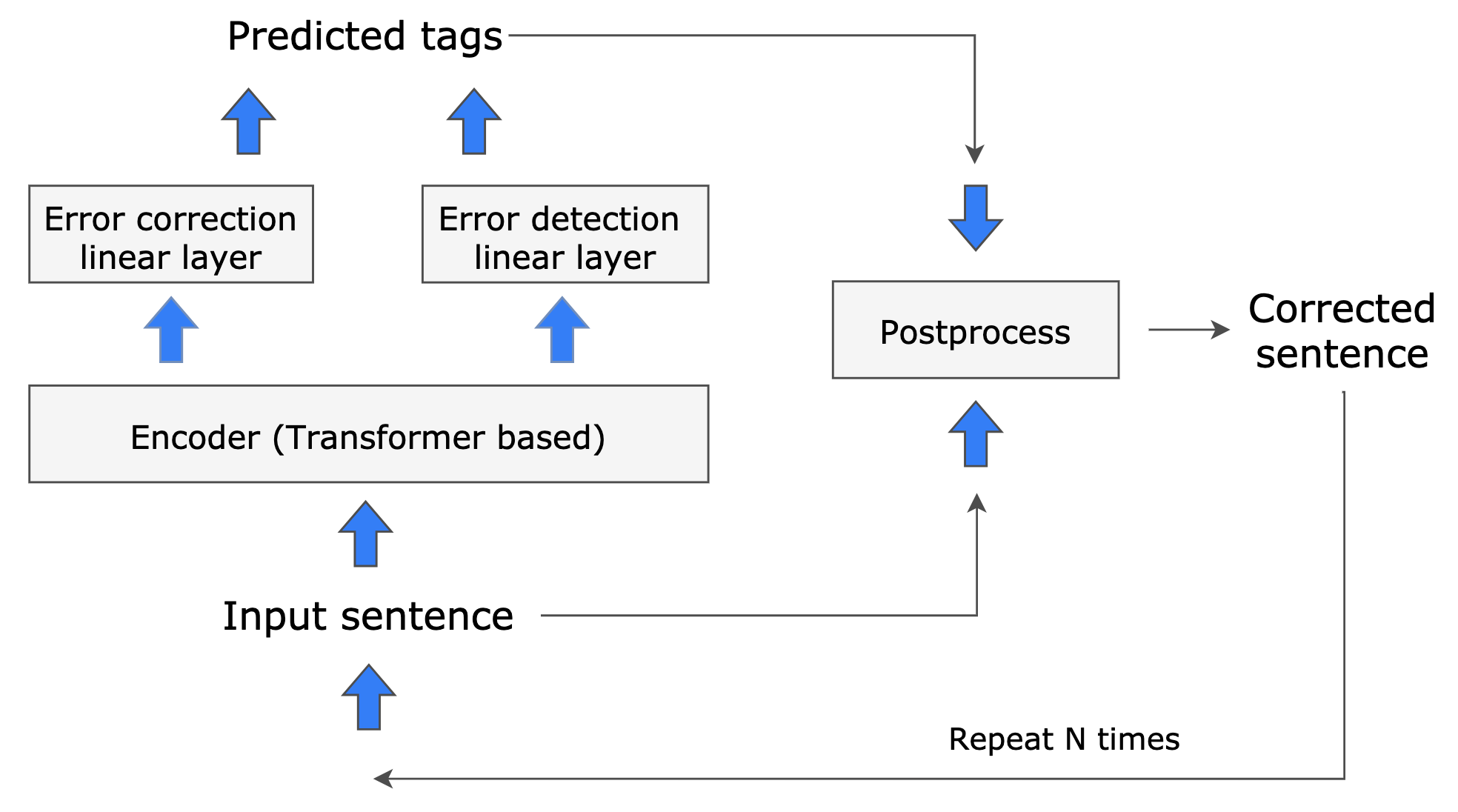} 
\caption{The GECToR model's iterative pipeline for sequence tagging and sentence modification.} 
\label{fig_gector}
\end{figure}

Since some corrections in a sentence may depend on others, applying the GEC sequence tagger only once may not be enough to correct the sentence entirely. Therefore, GECToR uses an iterative correction approach, modifying  the sentence by repeatedly running it through the model (up to four times) (Fig. \ref{fig_gector}).

\subsection{Tag-encoded edit operations} As in GECToR, our primary edit operations are encoded by the following tags: \textit{\$KEEP} (leave the current token unchanged), \textit{\$DELETE} (delete the current token), \textit{\$APPEND}\_$t_{1}$ (append the token $t_{1}$ after the current token), \textit{\$REPLACE}\_$t_{2}$  (replace the current token with the token $t_{2}$). GECToR also has special edit operations, such as changing the case of a token, changing the verb form to express a different number or tense, or converting singular nouns to plural, and other.  We refer to \cite{omelianchuk2020gector} for the details of edit transformations.

\subsection{Our contributions} 

We claim the following contributions:

1. We empirically investigate and improve the GECToR sequence tagging system \cite{omelianchuk2020gector} by upgrading the Transformer encoders to Large configurations, leveraging an advanced tokenizer, performing additional filtering of edits-free sentences, and increasing the vocabulary size.

2. We show that the ensembling of sequence taggers by majority votes on output edit spans provides better performance compared to ensembling by averaging output tag probabilities while staying tolerant to the models' architecture and vocabulary sizes.

3. We apply the knowledge distillation method to produce annotated data using ensemble of sequence taggers. When trained on the distilled data, single GEC tagging models show competitive performance.

4. We make the code, datasets, and trained models publicly available.

\section{Datasets}
 \begin{table}
 \scriptsize
 \begin{tabular}{|l|c|c|c|c|c|c|}
 \hline

 \textbf{Dataset} & \textbf{Type}  &  \textbf{Part} & \textbf{\# Sent.} & \textbf{\# Tokens} & \textbf{\% Edits} \\

  \hline
        Lang-8$^{\star}$  & Ann        & Train$^{\star}$          & 1.04M             & 11.86M & 42\%\\
 \hline
       NUCLE$^{\star}$    & Ann        & Train$^{\star}$          & 57k               & 1.16M & 62\% \\
 \hline
        FCE$^{\star}$      & Ann        & Train$^{\star}$          & 28k               & 455k & 62\% \\
  \hline
                &                   & Train$^{\star}$          & 34.3k             & 628.7k &  67\%  \\
         W\&I$^{\star\dagger}$   & Ann         & Dev            & 3.4k              & 63.9k &  69\% \\
                &                   & Test$^{\dagger}$           & 3.5k              & 62.5k &   N/A \\
  \hline

         LOCNESS$^{\dagger}$    & Ann         & Dev            & 1k              & 23.1k & 52\% \\
                &                   & Test$^{\dagger}$           & 1k              & 23.1k &   N/A\\
  \hline
        1BW$^{\ddag}$     & Mon       & N/A           &  115M  &  0.8B & N/A \\
 \hline        
        Blogs$^{\ddag}$     & Mon      & N/A           & 13.5M  & 171M & N/A \\
  \hline
        Troy-1BW     & Dis       & Train           &  1.2M  & 30.88M & 100\% \\
 \hline        
        Troy-Blogs     & Dis      & Train           & 1.2M  & 21.49M & 100\% \\
 \hline
        PIE$^{\ddag}$     & Syn        & Train           & 1.2M  & 30.1M & 100\%\\

\hline
\end{tabular}
\caption{Description and statistics of datasets used in this work. Dataset types: (Ann)otated, (Syn)thetic, (Mon)olingual, and (Dis)tilled. $^{\star}$Combined, these datasets form the \textit{Joint Train Dataset}. $^{\dagger}$BEA-2019 dev/test parts are concatenations of W\&I and LOCNESS dev/test parts. $^{\ddag}$Only parts of the original corpora from the cited sources are used in our work. }\label{training-data-table-new}
\end{table} 


\subsection{Annotated data} For training single models and ensembles, we use parallel annotated data from the Lang-8 Corpus of Learner English (Lang-8)\footnote{\url{https://sites.google.com/site/naistlang8corpora}} \cite{tajiri2012tense}, the National University of Singapore Corpus of Learner English (NUCLE)\footnote{\url{https://www.comp.nus.edu.sg/~nlp/corpora.html}} \cite{dahlmeier2013building}, the First Certificate in English dataset (FCE)\footnote{\url{https://ilexir.co.uk/datasets/index.html}} \cite{yannakoudakis2011new}, and the Write \& Improve (W\&I) Corpus \cite{bryant2019bea}.\footnote{\url{https://www.cl.cam.ac.uk/research/nl/bea2019st/data/wi+locness_v2.1.bea19.tar.gz}} Please, see Table \ref{training-data-table-new} for details.

\subsection{Monolingual data, distilled data} For knowledge distillation from the ensemble, we use parts of two monolingual datasets: the One Billion Word Benchmark (1BW)\footnote{\url{http://statmt.org/wmt11/training-monolingual.tgz}} \cite{one2013bw} and the Blog Authorship Corpus (Blogs)\footnote{\url{https://www.kaggle.com/rtatman/blog-authorship-corpus}} \cite{schl2005effec}. Corresponding distilled datasets have prefixes "Troy-"; see more details about their generation in Section 6.
\subsection{Synthetic data} 
After knowledge distillation for the final training of the student model we also use parallel sentences with synthetically generated grammatical errors from the PIE dataset \cite{awasthi-etal-2019-parallel}.\footnote{\url{https://github.com/awasthiabhijeet/PIE/tree/master/errorify}}
\subsection{Evaluation} We report $F_{0.5}$, $Precision$, and $Recall$  metrics computed by ERRANT scorer \cite{bryant2017automatic} on dev and test datasets from the W\&I + LOCNESS Corpus from the BEA-2019 GEC Shared Task \cite{bryant2019the}.

\section{Our System's Design}

\subsection{Tokenization} In the original GECToR system, the Byte-Pair Encoding (BPE) tokenizer \cite{sennrich-etal-2016-neural} uses a custom implementation.\footnote{\url{https://github.com/google/sentencepiece}} This was chosen because the out-off-the-box AllenNLP tokenizer was too slow, and HuggingFace Transformers' tokenizers did not provide a BPE-to-words mapping. Our work is fully implemented with Transformers from the HuggingFace Transformers library. In particular, we moved to the recently released fast tokenizers from HuggingFace. Now, our encoders have the same tokenizers for fine-tuning as they had for initial pretraining, which leads to better quality after fine-tuning.

\subsection{Initialization and training setup} Our encoder is loaded with its default pretrained weights; the linear layers' weights are initialized with random numbers. Our models are trained by Adam optimizer \cite{kingma1412adam} with default hyperparameters. We use a multi-class categorical cross-entropy loss function. The early stopping technique is used: Stopping criteria is 3 epochs without improving the loss function on the dev set, which is a random 2\% sample from the same source as training data and is different for each stage.

\subsection{Training stages} Model training is performed in several stages (Table \ref{train-stages}). In Stage I, the model is pretrained on synthetic datasets; this stage is optional. Then, in Stage II, we carry out warm-up training on the \textit{Joint Train Dataset}, which contains the Lang-8, NUCLE, FCE, and W\&I datasets (Table \ref{training-data-table-new}). Thus, we perform coarse fine-tuning on a large amount of diverse GEC data. Datasets are used sequentially with no shuffling. In order not to adversely impact the out-of-box pretrained weights of the encoder, during the first two epochs we train only the linear layers (so-called "cold epochs"); later, we make all model's weights trainable. 

In Stage III, we continue fine-tuning on the W\&I Train dataset, which contains only the highest-quality data. Another difference between Stages II and III is the share of edit-free sentences in the training data. We observed that too many sentences in training data without edits lead to reducing the appearance rate of the tagger and deteriorating the overall quality. Therefore, we filter out edit-free sentences from the Joint Train Dataset, which is used in Stage II. In Stage III, we fine-tune the model on the unfiltered version of the W\&I Train dataset. 

\begin{table}[htbp]
\scriptsize
\centering
\begin{tabular}{|l|ccc|ccc|}
\hline
\multicolumn{1}{|c|}{\multirow{2}{*}{\begin{tabular}[c]{@{}c@{}}\textbf{Training}\\ \textbf{stage} \#\end{tabular}}} & \multicolumn{3}{c|}{\textbf{Base}} & \multicolumn{3}{c|}{\textbf{Large}} \\ \cline{2-7} 
\multicolumn{1}{|c|}{}                                                                            & \textbf{P}           & \textbf{R}           & $\mathbf{F_{0.5}}$           & \textbf{P}          & \textbf{R}          & $\mathbf{F_{0.5}}$          
\\ \hline
Stage I.  &   N/A & N/A	 & N/A   &   N/A & N/A & N/A   \\ 
Stage II.  &   50.12 & 34.04	 & 45.79    &   52.11 & 37.34 & 48.29   \\ 
Stage III.   &  53.77 & \textbf{39.23}& 50.06      &   54.85 & \textbf{42.54} & 51.85   \\ 
Inf. tweaks &\textbf{62.49} & 32.26	 & \textbf{52.63}  & \textbf{65.76} & 33.86 &\textbf{55.33}   \\

\hline
\end{tabular}
\caption{\label{train-stages} Performance of our system with a RoBERTa encoder (in Base and Large configurations) after each training stage and inference tweaks on BEA-2019 (dev). Pre-training on synthetic data (Stage I) as was done in \cite{omelianchuk2020gector} is not performed.}
\end{table}

The final stage is inference tweaks \cite{omelianchuk2020gector} for balancing between the model's precision and recall. This is done by introducing additional hyperparameters: additional confidence (AC) to the probability for the \textit{\$KEEP} tag and minimum error probability (MEP) for corrections tags. These hyperparameters are found via a random search on the BEA-2019 dev set.

\subsection{Upgrading to Large encoders} In the GECToR paper \cite{omelianchuk2020gector}, authors investigated encoders from ALBERT \cite{lan2020albert}, BERT \cite{devlin2018bert}, GPT-2 \cite{gpt-2}, RoBERTa \cite{liu2019roberta}, and XLNet \cite{yang2019xlnet} Transformers in their Base configurations. Most likely, Base configurations were chosen due to the better inference speed/quality ratio. They found that XLNet, RoBERTa, and BERT show the best performance. 

We reproduce experiments for these encoders, but now we explore Large configurations as well. We additionally explore encoders from DeBERTa \cite{he2020deberta} (Table \ref{table-comparison-encoder-types}). 

\begin{table}[htbp]
\scriptsize
\centering
\begin{tabular}{|l|ccc|ccc|}
\hline
\multicolumn{1}{|c|}{\multirow{2}{*}{\begin{tabular}[c]{@{}c@{}}\textbf{Encoder}\\\end{tabular}}} & \multicolumn{3}{c|}{\textbf{Base}} & \multicolumn{3}{c|}{\textbf{Large}} \\ \cline{2-7} 
\multicolumn{1}{|c|}{} & \textbf{P} & \textbf{R}  & $\mathbf{F_{0.5}}$  & \textbf{P} & \textbf{R} & $\mathbf{F_{0.5}}$  \\ \hline
BERT & 57.21 & 29.93 & 48.39  & 61.18 & 31.26	& 51.35  \\
DeBERTa   & \textbf{64.22} & 31.87 &\textbf{53.38}    &   \textbf{66.35} & 32.77	& 55.07    \\ 
RoBERTa   & 62.49 & \textbf{32.26}	& 52.63  & 65.76 & 33.86 &\textbf{55.33}   \\ 
XLNet & 63.16 & 30.59 & 52.07  &  64.27 &\textbf{35.17}	& 55.14   \\

\hline
\end{tabular}
\caption{\label{table-comparison-encoder-types} Performance of our single system on BEA-2019 (dev) for different encoders from pretrained Transformers in Base and Large configurations.}
\end{table}

\begin{table}[htbp]
\scriptsize
\begin{tabular}{|l|c|c|c|c|}
\hline
\multirow{2}{*}{\textbf{Encoder}} & \multicolumn{2}{l|}{\textbf{Time (sec)}} & \multicolumn{2}{l|}{\textbf{\# Params}} \\ \cline{2-5} 
                                  & \textbf{Base}      & \textbf{Large}     & \textbf{Base}     & \textbf{Large}     \\ \hline
BERT                              & 19.28              & 49.17              & 120M              & 350M               \\ \hline
DeBERTa                           & 23.75              & 58.32              & 150M              & 410M               \\ \hline
RoBERTa                           & 19.05              & 45.66              & 129M              & 360M               \\ \hline
XLNet                             & 30.46              & 71.19              & 120M              & 345M               \\ \hline
\end{tabular}
\caption{\label{table-comparison-encoder-types-time} Inference times and model sizes for our single tagging models. Inference time for NVIDIA Tesla P100 on BEA-2019 dev, single models, batch size=128, averaged over 5 inferences.}
\end{table}



We observe that all models that are equipped with Large encoders have higher precision, recall, and $F_{0.5}$ values than those equipped with their Base versions. The price of this performance is 2.3–2.5 times slower inference for Large configurations (Table \ref{table-comparison-encoder-types-time}). The single model with RoBERTa encoder shows the best performance for Large configurations, whereas DeBERTa slightly outperforms RoBERTa for Base configurations. RoBERTa is the fastest in both configurations.

\subsection{Exploring tag vocabulary sizes} Most of the tag-encoded edits are token-specific, e.g., \textit{\$APPEND\_it}, \textit{\$REPLACE\_the}, and so on. Thus, the tag vocabulary size matters, and should be a tradeoff between coverage and model quality.

We create the tag vocabulary by taking the most frequent edit tags generated from the Joint Train Dataset (Table \ref{training-data-table-new}). To find the optimal tag vocabulary sizes, we experiment with \{5K, 10K\} vocabulary sizes (Table \ref{table-vocab-sizes}). 

\begin{table}[htbp]
\centering
\footnotesize
\begin{tabular}{l c c c }
    \toprule
    \textbf{Encoder} & {$\mathbf{P}$} & {$\mathbf{R}$} & {$\mathbf{F_{0.5}}$}        \\ \midrule
    
        
    DeBERTa$^{(L)}_{5K}$ & \textbf{66.35} & 32.77 & 55.07\\
    RoBERTa$^{(L)}_{5K}$ & 65.76 & 33.86 &\textbf{55.33}\\
    XLNet$^{(L)}_{5K}$   & 64.27 & \textbf{35.17} & 55.14\\ \midrule

    DeBERTa$^{(L)}_{10K}$ & \textbf{65.46} & 34.59 & 55.55\\
    RoBERTa$^{(L)}_{10K}$ & 64.72 & \textbf{36.04} &\textbf{55.83}\\
    XLNet$^{(L)}_{10K}$   & 64.12 & 34.02 & 54.48      \\\bottomrule
\end{tabular}
\caption{Performance on BEA-2019 (dev) for varied tag vocabulary sizes and encoders in their (L)arge configurations. Subscripts encode the models' tag vocabulary sizes from the set (5K, 10K). 
}\label{table-vocab-sizes}
\end{table}

We observe that increasing the vocabulary size to 10K for Large encoders may improve the quality, e.g. for models with RoBERTa and DeBERTa. Nevertheless, we also see an example of quality deterioration for the model with XLNet.

\section{Ensembling the GEC taggers}
\begin{figure}[!h]
\centering
\includegraphics[width=1.0\columnwidth]{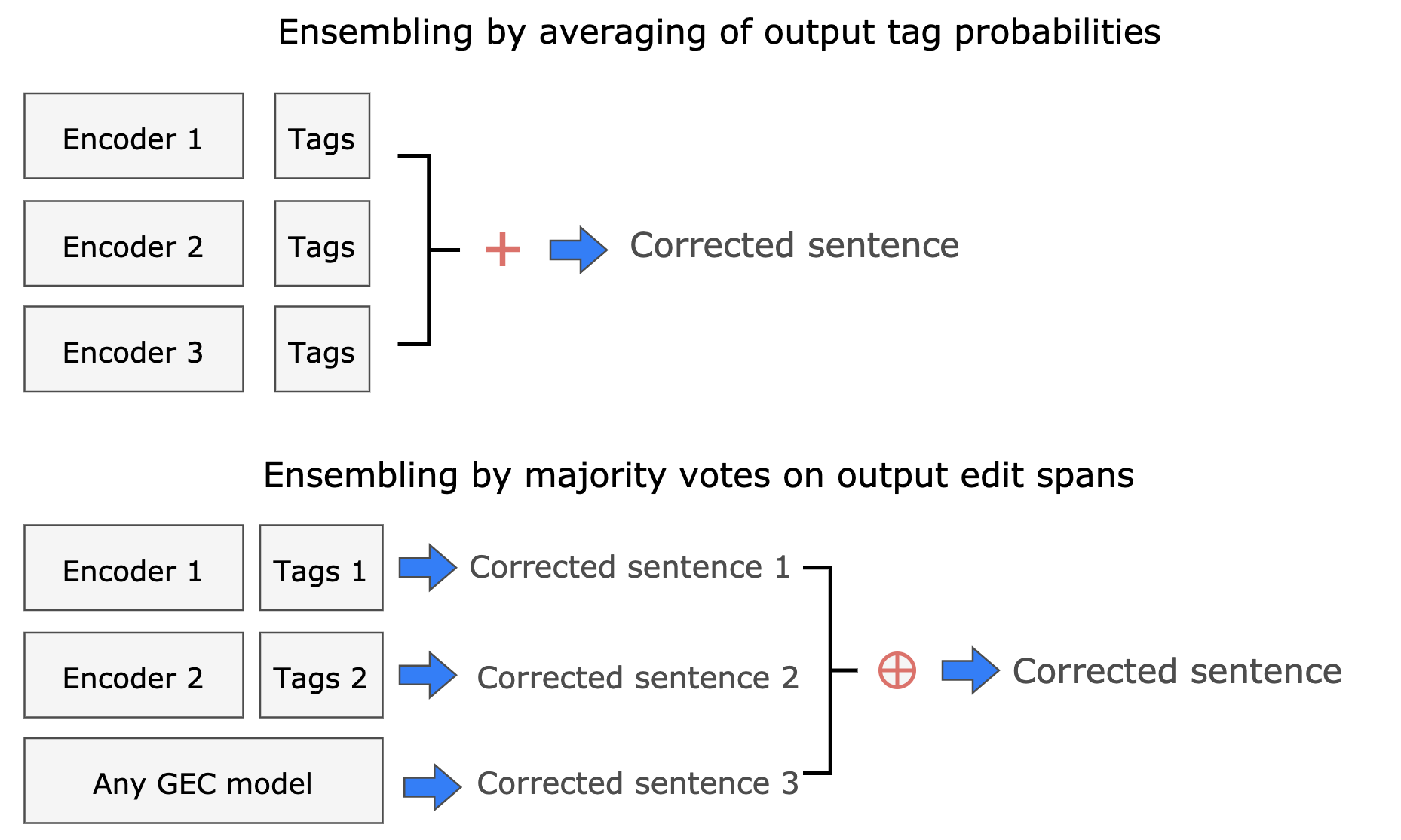} 
\caption{Ensembling by averaging of output tag probabilities (top) and ensembling by majority votes on output edit spans (bottom).} 
\label{fig_ens}
\end{figure}

Ensembling is a proven quality-boosting method for models sets that have diverse outputs. Most of the recent GEC solutions achieved their best results by ensembling single models \cite{stahlberg2021synthetic}, \cite{omelianchuk2020gector}, \cite{awasthi2019parallel}. In this section we consider two ensembling methods for our GEC tagging models: averaging of output tag probabilities and majority votes on output edit spans (Fig. \ref{fig_ens}). 

\subsection{Exploring averaging of output tag probabilities (``+'' operation)} First, we reproduce the ensembling approach from \cite{omelianchuk2020gector}. We add DeBERTa and carry out experiments with varying Base and Large configurations of encoders (Table \ref{table-ens-average}). 

\begin{table}[htbp]
\centering
\scriptsize
\begin{tabular}{l c c c }
    \toprule
    \textbf{Ensemble} & {$\mathbf{P}$} & {$\mathbf{R}$} & {$\mathbf{F_{0.5}}$}        \\ \midrule
    RoBERTa$^{(B)}$ + DeBERTa$^{(B)}$  & 53.44 &\textbf{34.91} & 48.31\\
    RoBERTa$^{(B)}$ + XLNet$^{(B)}$    & 53.45 & 34.3 & 48.08      \\
    RoBERTa$^{(B)}$ + DeBERTa$^{(B)}$ + XLNet$^{(B)}$  & 54.78 & 34.87 & 49.17\\
    RoBERTa$^{(B)}$ + BERT$^{(B)}$ + DeBERTa$^{(B)}$ +   &  &   & \\
     + XLNet$^{(B)}$  & \textbf{56.34} & 33.76 &\textbf{49.69}\\ \midrule  
    RoBERTa$^{(B)}$    &  50.12 &  34.04 & 45.79 \\
    RoBERTa$^{(L)}$    & 52.11 & \textbf{37.34} & 48.29      \\
    RoBERTa$^{(B)}$ + RoBERTa$^{(L)}$    & \textbf{54.83} & 35.93 & \textbf{49.61}      \\ \midrule 
    RoBERTa$^{(L)}$ + DeBERTa$^{(L)}$    & 54.12 & 39.77 & 50.48      \\
    RoBERTa$^{(L)}$ + XLNet$^{(L)}$    & 53.83 & 38.65 & 49.91      \\
    RoBERTa$^{(L)}$ + BERT$^{(L)}$ + DeBERTa$^{(L)}$  & \textbf{57.31} & 37.41 & 51.8\\
    RoBERTa$^{(L)}$ + DeBERTa$^{(L)}$ + XLNet$^{(L)}$  & 54.30 & \textbf{39.95} & 50.66\\
    RoBERTa$^{(L)}$ + BERT$^{(L)}$ + DeBERTa$^{(L)}$ +  & &  &       \\
     + XLNet$^{(L)}$    & 56.97 & 38.52 & \textbf{51.99}       \\

    \bottomrule    
\end{tabular}
\caption{Comparison of ensembles by averaging of output tag probabilities after Stage II for (B)ase and (L)arge encoders with a tag vocabulary size of 5K. Benchmark is BEA-2019 (dev).
}\label{table-ens-average}
\end{table} 

We observe that ensembling by averaging of output tag probabilities improves the quality of corrections; the more models we combine, the better results we obtain. More surprisingly, combining the same encoders' architectures in Base and Large configurations may provide slightly better results than we get for the Base and Large models separately (see RoBERTa$^{(B)}$ + RoBERTa$^{(L)}$ in Table \ref{table-ens-average}).

Although the ensemble RoBERTa$^{(L)}$ + BERT$^{(L)}$ + DeBERTa$^{(L)}$ + XLNet$^{(L)}$ shows the best performance, we select ensemble the RoBERTa$^{(L)}$ + DeBERTa$^{(L)}$ + XLNet$^{(L)}$ for further experiments. It has higher recall, making it possible to trade recall for precision later during inference tweaks.

\subsection{Exploring majority votes on output edit spans (``$\mathbf{\oplus}$'' operation)} This aggregation method combines single models' outputs in the post-processing step (Fig. \ref{fig_ens}). We take span-level edits and retain only those which have most of the votes from the ensemble. A similar approach is used in \cite{liang2020bert}, where the authors combined sequence tagging and seq2seq models for the Chinese language. The advantage of this ensembling method is that we can combine the results of models with different output dimensions and even different architectures. In our work, it allows us to combine models with different tag vocabulary sizes. We leave ensembling with seq2seq GEC systems for future work.

First, we compare ensembling by averaging of output tag probabilities $"+"$ and by majority votes on output edit spans $\oplus$ for the selected ensemble after training on the Joint Train Dataset (Stage II), finetuning on the W\&I dataset (Stage III) and optimization of hyperparameters (inference tweaks) (Table \ref{table-ens-mv-stages}). We observe that ensembles based on majority votes on output edit spans show better results because of better precision. However, after inference tweaks, the two ensembling types achieve close $F_{0.5}$ scores.

\begin{table}[htbp]
\centering
\tiny
\begin{tabular}{l l c c c }
    \toprule
    \textbf{Stage} & \textbf{Ensemble} & {$\mathbf{P}$} & {$\mathbf{R}$} & {$\mathbf{F_{0.5}}$}        \\ 
    \midrule
    St. I & RoBERTa$^{(L)}$ + DeBERTa$^{(L)}$ + XLNet$^{(L)}$    & N/A & N/A & N/A      \\
    St. I & RoBERTa$^{(L)}$ $\oplus$ DeBERTa$^{(L)}$ $\oplus$ XLNet$^{(L)}$   & N/A & N/A & N/A \\
    \midrule
    St. II & RoBERTa$^{(L)}$ + DeBERTa$^{(L)}$ + XLNet$^{(L)}$    & 54.3 & \textbf{39.95} & 50.66      \\
    St. II & RoBERTa$^{(L)}$ $\oplus$ DeBERTa$^{(L)}$ $\oplus$ XLNet$^{(L)}$   & \textbf{56.74} & 38.53 & \textbf{51.84} \\
    \midrule
    St. III & RoBERTa$^{(L)}$ + DeBERTa$^{(L)}$ + XLNet$^{(L)}$    & 58.08 & \textbf{43.17} & 54.33      \\
    St. III & RoBERTa$^{(L)}$ $\oplus$ DeBERTa$^{(L)}$ $\oplus$ XLNet$^{(L)}$  &\textbf{60.58} & 41.92 & \textbf{55.63} \\
    \midrule
    In.tw. & RoBERTa$^{(L)}$ + DeBERTa$^{(L)}$ + XLNet$^{(L)}$ & 68.45 & \textbf{35.56} & 57.76      \\
    In.tw. & RoBERTa$^{(L)}$ $\oplus$ DeBERTa$^{(L)}$ $\oplus$ XLNet$^{(L)}$  & \textbf{69.67} & 34.51 & \textbf{57.88}\\
    \bottomrule     
\end{tabular}
\caption{Performance of selected ensemble for averaging of output tag probabilities ("+") and majority votes on output edit spans ("$\oplus$") ensembling types. Ensembles are not pre-trained on synthetic data (Stage I), tag vocabulary size of 5K. Benchmark is BEA-2019 (dev). }\label{table-ens-mv-stages}
\end{table}

To additionally improve the precision of ensembling by majority votes we introduce the "majority quorum" hyperparameter $N_{min}$. Majority quorum $N_{min}$ denotes \textit{minumum number of votes for triggering the edit}, here $1 \leq N_{min} \leq N_{single\_models}$. Increasing $N_{min}$ boosts precision by the cost of recall because it filters out more edits where single models disagree (Table \ref{table-ens-maj-votes-mink}). Setting $N_{min} = 1$ is a poor strategy because we can't rely on a majority when resolving conflicting edits, so the resulting text might contain controversial and incoherent edits.

Increasing the number of systems in the ensemble leads to higher quality, but requires adapting the $N_{min}$ parameter (Table \ref{table-ens-maj-votes-mink}). Based on this limited analysis we observe that $N_{min}$ = $N_{single\_models} - 1$  achieves the best results. For our pool of models there is no gain over using more than 4 models, but we want to explore adding more diverse seq2seq models to such an ensemble in future works. 

\begin{table*}[ht]
\centering
\scriptsize
\begin{tabular}{l c c c c c}
    \toprule
     \textbf{Ensemble} & $\mathbf{N_{single\_models}}$ & $\mathbf{N_{min}}$  & {$\mathbf{P}$} & {$\mathbf{R}$} & {$\mathbf{F_{0.5}}$} \\ 
    \midrule
    RoBERTa$^{(B)}_{5K}$ $\oplus$ RoBERTa$^{(L)}_{5K}$ $\oplus$ RoBERTa$^{(L)}_{10K}$ & 3 & 1 & 44.49 & \textbf{41.96} & 43.96 \\
    RoBERTa$^{(B)}_{5K}$ $\oplus$ RoBERTa$^{(L)}_{5K}$ $\oplus$ RoBERTa$^{(L)}_{10K}$ & 3 & 2 & 57.96 & 41.79 & 53.79 \\
    RoBERTa$^{(B)}_{5K}$ $\oplus$ RoBERTa$^{(L)}_{5K}$ $\oplus$ RoBERTa$^{(L)}_{10K}$ & 3 & 3 & \textbf{67.54} & 30.99 & \textbf{54.65} \\\midrule
    RoBERTa$^{(B)}_{5K}$ $\oplus$ RoBERTa$^{(L)}_{5K}$ $\oplus$ RoBERTa$^{(L)}_{10K}$  $\oplus$ DeBERTa$^{(L)}_{10K}$ & 4 & 1 & 40.21 & 41.68 & 40.50 \\
    RoBERTa$^{(B)}_{5K}$ $\oplus$ RoBERTa$^{(L)}_{5K}$ $\oplus$ RoBERTa$^{(L)}_{10K}$  $\oplus$ DeBERTa$^{(L)}_{10K}$ & 4 & 2 & 55.02 & \textbf{43.14} & 52.15 \\
    RoBERTa$^{(B)}_{5K}$ $\oplus$ RoBERTa$^{(L)}_{5K}$ $\oplus$ RoBERTa$^{(L)}_{10K}$  $\oplus$ DeBERTa$^{(L)}_{10K}$ & 4 & 3 & 64.48 & 37.49 & \textbf{56.36} \\
    RoBERTa$^{(B)}_{5K}$ $\oplus$ RoBERTa$^{(L)}_{5K}$ $\oplus$ RoBERTa$^{(L)}_{10K}$  $\oplus$ DeBERTa$^{(L)}_{10K}$ & 4 & 4 & \textbf{71.71} & 27.89 & 54.57 \\\midrule
    RoBERTa$^{(B)}_{5K}$ $\oplus$ RoBERTa$^{(L)}_{5K}$ $\oplus$ RoBERTa$^{(L)}_{10K}$  $\oplus$ DeBERTa$^{(L)}_{10K}$  $\oplus$ XLNet$^{(L)}_{10K}$ & 5 & 1 & 37.20 & 40.88 & 37.88 \\
    RoBERTa$^{(B)}_{5K}$ $\oplus$ RoBERTa$^{(L)}_{5K}$ $\oplus$ RoBERTa$^{(L)}_{10K}$  $\oplus$ DeBERTa$^{(L)}_{10K}$  $\oplus$ XLNet$^{(L)}_{10K}$ & 5 & 2 & 51.77 & \textbf{43.65} & 49.92 \\
    RoBERTa$^{(B)}_{5K}$ $\oplus$ RoBERTa$^{(L)}_{5K}$ $\oplus$ RoBERTa$^{(L)}_{10K}$  $\oplus$ DeBERTa$^{(L)}_{10K}$  $\oplus$ XLNet$^{(L)}_{10K}$ & 5 & 3 & 61.89 & 41.43 & 56.33 \\
    RoBERTa$^{(B)}_{5K}$ $\oplus$ RoBERTa$^{(L)}_{5K}$ $\oplus$ RoBERTa$^{(L)}_{10K}$  $\oplus$ DeBERTa$^{(L)}_{10K}$  $\oplus$ XLNet$^{(L)}_{10K}$ & 5 & 4 & 56.43 & 34.43 & \textbf{56.43} \\
    RoBERTa$^{(B)}_{5K}$ $\oplus$ RoBERTa$^{(L)}_{5K}$ $\oplus$ RoBERTa$^{(L)}_{10K}$  $\oplus$ DeBERTa$^{(L)}_{10K}$  $\oplus$ XLNet$^{(L)}_{10K}$ & 5 & 5 & \textbf{73.12} & 26.00 & 53.67 \\
    \bottomrule     
\end{tabular}
\caption{Exploring the impact of $N_{min}$ ("majority quorum"), a minumum number of votes to trigger the edit in majority votes ensembling. Benchmark is BEA-2019 (dev).}
\label{table-ens-maj-votes-mink}
\end{table*}

Next, since the majority votes on output edit spans is capable of combining any models, we test the ensemble of the best models that we already have trained (Table \ref{table-ens-best}).

\begin{table}[htbp]
\centering
\scriptsize
\begin{tabular}{p{0.29\textwidth} c c c }
    \toprule
    \textbf{Ensemble} & {$\mathbf{P}$} & {$\mathbf{R}$} & {$\mathbf{F_{0.5}}$}        \\ \midrule
    DeBERTa$^{(L)}_{5K}\oplus$RoBERTa$^{(L)}_{5K}\oplus$XLNet$^{(L)}_{5K}$ & 69.67 & 34.51 & 57.88      \\
    DeBERTa$^{(L)}_{10K}\oplus$RoBERTa$^{(L)}_{10K}\oplus$XLNet$^{(L)}_{10K}$ & 70.13 & 34.23 & 57.97      \\
    DeBERTa$^{(L)}_{5K}\oplus$RoBERTa$^{(L)}_{10K}\oplus$XLNet$^{(L)}_{5K}$ &\textbf{70.71} & 33.78 & 58.02      \\
    DeBERTa$^{(L)}_{10K}\oplus$RoBERTa$^{(L)}_{10K}\oplus$XLNet$^{(L)}_{5K}$ & 70.32 &\textbf{34.62} & \textbf{58.30}\\\bottomrule
\end{tabular}
\caption{ 
Performance of the best single models ensembled by majority votes on output edit spans. Subscripts encode the models' tag vocabulary sizes from the set (5K, 10K). 
Benchmark is BEA-2019 (dev).
}
\label{table-ens-best}
\end{table}

Finally, we evaluate our best ensemble DeBERTa$^{(L)}_{10K}$ $\oplus$ RoBERTa$^{(L)}_{10K}$ $\oplus$ XLNet$^{(L)}_{5K}$ on the BEA-2019 (test) dataset and achieve $F_{0.5}$ score of $76.05$. This is a significant improvement over $F_{0.5}=73.70$ for the best ensemble from \cite{omelianchuk2020gector} and to the best of our knowledge \textbf{is a new state-of-the-art (SOTA) result for ensembles on the BEA-2019 (test) benchmark}. It is worth noting that the solution is obtained without pre-training on synthetic data.

\section{Knowledge distillation}

Knowledge distillation is the method for transferring knowledge from a large model ("teacher") to a smaller one ("student") \cite{hinton2015distilling}, \cite{kim-rush-2016-sequence}. It has strong practical applications because large models usually have expensive inference costs and are inconvenient for deployment. 

In our case, the teacher model is an ensemble of trained sequence taggers, whereas the student model is a single sequence tagger. The ensemble receives errorful texts and generates their corrected versions. Later these input-output pairs of sentences are used for training single models. Like any synthetic annotation method, knowledge-distilled data contains a certain share of systematic errors that deteriorates the student model's quality.
\subsection{Distilling the data} In this work, we use two monolingual corpora to generate our distilled datasets: the One Billion Words Benchmark ("1BW"), which mostly contains news texts, and the Blog Authorship Corpus ("Blogs"), which contains blog texts on various topics (Table \ref{training-data-table-new}). Being real-world natural texts, these datasets contain a certain share of grammatical errors, which are corrected by our system. For text pre-processing, we use the tokenizer from Spacy.\footnote{ \url{https://spacy.io/}}

As a teacher, we use the ensemble of the sequence taggers containing Large encoders with a 5K vocabulary:  DeBERTa$^{(L)}_{5K}$ + RoBERTa$^{(L)}_{5K}$ + XLNet$^{(L)}_{5K}$ (Table \ref{table-ens-mv-stages}). The ensemble corrects 5\% of processed sentences in 1BW and 28\% of sentences in Blogs. Distilled versions of the datasets have the prefix "Troy-" in their names (Table \ref{training-data-table-new}). Considering our past experience, we use only edited sentence pairs in our distilled datasets, and we limit their number to 1.2M. We also reduce the synthetic PIE dataset from \cite{awasthi-etal-2019-parallel} to 1.2M sentence pairs for better comparability in the experiments. We leave exploring other ensembles in the role of a teacher model for future research. 

\subsection{Pre-training on synthetic and distilled datasets ("multi-stage training")} 
\begin{figure}[!h]
\centering
\includegraphics[width=1.0\columnwidth]{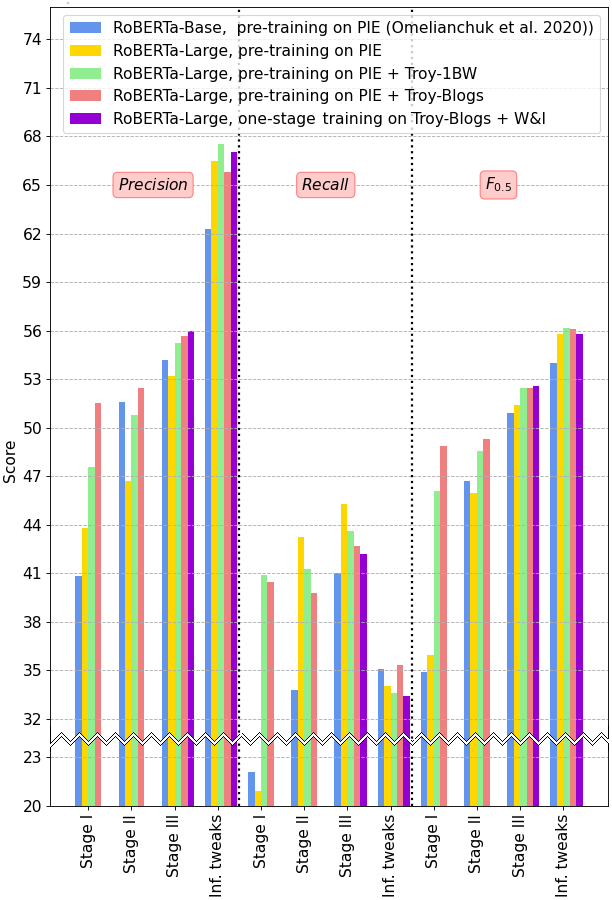} 
\caption{Pre-training of single tagging models on synthetic and distilled datasets with a tag vocabulary size of 5K. Benchmark is BEA-2019 (dev).} 
\label{fig_dist}
\end{figure}

First, we reproduce the training scheme from \cite{omelianchuk2020gector} for a single model, RoBERTa$^{(L)}_{5K}$ where PIE synthetic data is used for pre-training (Stage I), then the model is trained on the Joint Train Dataset (Stage II), fine-tuned on the high-quality W\&I dataset (Stage III), and finally, hyperparameters are applied to balance precision and recall (inteference tweaks). We observe that the sequence tagger with a RoBERTa-Large encoder shows slightly better performance than RoBERTa-Base from \cite{omelianchuk2020gector}, where RoBERTa-Base had an 8x larger training dataset in Stage I (Fig. \ref{fig_dist}). 

Next, we replace the synthetic PIE dataset with our distilled datasets, Troy-1BW and Troy-Blogs. We observe that in Stage I, training on purely synthetic data leads to a dramatic boost in recall. When we start training in Stage II, a sharp deterioration in both precision and recall occurs. It seems that the student model does not receive new information compared to Stage I. This is more noticeable for models trained on the Troy-Blogs dataset, where recall significantly drops after training. However, the $F_{0.5}$ in Stage II is higher for models pretrained on distilled Troy- datasets. 

Finally, after training on Stage III and performing inference tweaks, single models pretrained on both datasets show very similar performance, but the model with RoBERTa$^{(L)}_{5K}$ trained on Troy-1BW is slightly higher-performing. \textbf{This single model reaches $\mathbf{F_{0.5} = 73.21}$ on BEA-2019 (test), a significant improvement on the results from} \cite{omelianchuk2020gector} \textbf{for single models}  $F_{0.5} = 71.5$ for RoBERTa$^{(B)}_{5K}$ and $F_{0.5} = 72.4$ for XLNet$^{(B)}_{5K}$.

\begin{table}[htb]
\scriptsize
\centering
\begin{tabular}{lccc}
\hline
\textbf{System} & \textbf{P}  & \textbf{R}  &  $\mathbf{F_{0.5}}$   \\
\hline
\textbf{Single models} &  &  &   \\
\hline
\cite{kiyono2019empirical} & 65.5 & 59.4 & 64.2  \\
\cite{omelianchuk2020gector} & 79.2 & 53.9 & 72.4  \\
\cite{kaneko2020encoder} & 67.1 & 60.1 &  65.6  \\
\cite{stahlberg2021synthetic} & 72.1 & \textbf{64.4} & 70.4 \\
\cite{rothe2021a} & N/A & N/A & \textbf{75.88} \\
RoBERTa$^{(L)}_{5K}$, multi-stage training (this work) & \textbf{80.70} & 53.39 & 73.21 \\
RoBERTa$^{(L)}_{5K}$, one-stage training (this work) & 80.55 & 52.27 & 72.69 \\
\hline
\textbf{Ensembles} &  &  &  \\
\hline
\cite{grundkiewicz-etal-2019-neural} & 72.3 & 60.1 & 69.5 \\
\cite{kiyono2019empirical} & 74.7 & 56.7 & 70.2 \\
\cite{omelianchuk2020gector} & 79.4 & 57.2 & 73.7  \\
\cite{kaneko2020encoder} & 72.3 & 61.4 & 69.8  \\
\cite{stahlberg2021synthetic} & 77.7 & \textbf{65.4} & 74.9 \\
DeBERTa$_{10K}^{(L)}$ $\oplus$ RoBERTa$_{10K}^{(L)}$ $\oplus$ XLNet$_{5K}^{(L)}$ & \textbf{84.44} & 54.42 & \textbf{76.05} \\
 (this work) &   &  &  \\
\hline
\end{tabular}
\caption{\label{table_related} Comparison of our best single tagging models and ensembles with related work on BEA-2019 (test).}
\end{table}

\subsection{One-stage training on distilled + annotated dataset} We observed that models pretrained on the Troy-Blogs dataset show good results on Stage I, but lose their advantage after training on Stage II. Thus, we decided to try a one-stage training approach with a RoBERTa$^{(L)}_{5K}$ encoder. 

For our training dataset, we concatenated Troy-Blogs with high-quality W\&I dataset that we usually reserve for Stage III. As a result, we achieved $F_{0.5} = 55.81$ on BEA-2019 (dev) and $F_{0.5} = 72.69$ on BEA-2019 (test) (Table \ref{table_related}). These results \textbf{are obtained much more easily than with our best single model: just one-stage training with out-of-the-box RoBERTa}, no pre-training on synthetic GEC data or multi-stage training. 

\section{Conclusions}
Our research investigates the impact of encoder configurations, ensembling methods, and knowledge distillation on the GECToR system.

We found that Replacing Base encoders in GECToR \cite{omelianchuk2020gector} with their Large configurations does improve the quality by several F0.5 points, at the cost of 2.3–2.5 times slower inference. 

Our best ensemble achieves a new SOTA result with $F_{0.5} = 76.05$ on BEA-2019 (test). Ensembling sequence taggers by majority votes on output edit spans provides better performance than averaging output tag probabilities because it lets us combine a variety of modeling approaches and vocabulary sizes. Single models in the ensemble were not pre-trained on synthetic GEC datasets, providing room for improvement in future work. 

We apply the knowledge distillation method to an ensemble of sequence taggers and produce the annotated Troy-Blogs and Troy-1BW datasets. After training on these datasets, single GEC sequence tagging models show near-SOTA results: $F_{0.5} = 73.21/72.69$ on BEA-2019 (test) for multi-stage/one-stage training. To our knowledge, our best single model is outperformed only by the much more compute-intensive T5 XXL model \cite{rothe2021a}, which is 30 times larger with 11B parameters (Table \ref{table_related}). 

We make the code, datasets, and trained models publicly available.\footnote{\url{https://github.com/MaksTarnavskyi/gector-large}}

\section{Acknowledgements}
We express our gratitude to Oleksii Molchanovskyi, Dmytro Lider, Viktor Zamaruiev, Paige Schwartz, the Ukrainian Catholic University, and Grammarly for providing support and computational resources. We also thank anonymous reviewers for their contributions. To our communities: While we are writing this, our homeland Ukraine continues to resist the unprovoked Russian invasion. We are grateful to everyone who defends Ukraine, declares support to the people of Ukraine, and is sending aid. Thank you!

\bibliography{main}
\bibliographystyle{main}

\appendix

\newpage

\section{Appendix}
\label{sec:appendix}

\begin{table}[ht]
\scriptsize
\centering
\begin{tabular}{lccc}
\hline
\textbf{System} & \textbf{P}  & \textbf{R}  &  $\mathbf{F_{0.5}}$   \\
\hline
\textbf{Single models} &  &  &   \\
\hline
\cite{kiyono2019empirical} & 67.9 & 44.1 & 61.3  \\
\cite{omelianchuk2020gector} & \textbf{77.5} & 40.1 & 65.3  \\
\cite{kaneko2020encoder} & 69.2 & 45.6 &  62.6  \\
\cite{stahlberg2021synthetic} & 72.8 & \textbf{49.5} & 66.6 \\
\cite{rothe2021a} & N/A & N/A & \textbf{68.9} \\
RoBERTa$^{(L)}_{5K}$, multi-stage training (this work) & 74.40 & 41.05 & 64.0 \\
RoBERTa$^{(L)}_{5K}$, one-stage training (this work) & 70.12 & 42.66 & 62.12 \\
\hline
\textbf{Ensembles} &  &  &  \\
\hline
\cite{grundkiewicz-etal-2019-neural} & N/A & N/A &  64.2 \\
\cite{kiyono2019empirical} & 72.4 & 46.1 & 65.0 \\
\cite{omelianchuk2020gector} & \textbf{78.2} & 41.5 & 66.5  \\
\cite{kaneko2020encoder} & 72.6 &  46.4 &  65.2  \\
\cite{stahlberg2021synthetic} & 75.6 & \textbf{49.3} & \textbf{68.3} \\
DeBERTa$_{10K}^{(L)}$ $\oplus$ RoBERTa$_{10K}^{(L)}$ $\oplus$ XLNet$_{5K}^{(L)}$
 (this work) & 76.1 & 41.6 & 65.3 \\
\hline
\end{tabular}
\caption{\label{table_related_nucle} Comparison of our best single tagging models and ensembles with related work on CoNLL-14 (test).}
\end{table}

\end{document}